\documentclass[runningheads]{llncs}
\usepackage[T1]{fontenc}
\usepackage{graphicx}
\usepackage{booktabs}
\usepackage[misc]{ifsym}
\newcommand{\corr}{(\Letter)}
\usepackage{amssymb}
\usepackage{mathtools}
\usepackage{lipsum} 
\usepackage{siunitx}
\usepackage{upgreek}
\usepackage[noend]{algpseudocode}
\usepackage{algorithm}
\usepackage{multirow}
\usepackage{anyfontsize}
\usepackage{diagbox}
\usepackage{url}
\usepackage{tabularx,booktabs}
\usepackage{tabto}
\usepackage{breqn}
\usepackage{bbm}
\usepackage{enumitem}
\usepackage{balance}
\usepackage{upgreek}
\usepackage{mathrsfs}
\usepackage{multirow}
\usepackage{longtable}
\usepackage{lscape}
\usepackage{makecell}
\usepackage{hvfloat}
\usepackage{hyperref}
\usepackage[table]{xcolor}

\usepackage{comment}

\usepackage{chngcntr}
\counterwithin{figure}{section}
\counterwithin{table}{section}

\begin{document}

\title{Procedural Fairness via Group Counterfactual Explanation}

\titlerunning{Group Counterfactual Explanation}




\author{Gideon Popoola \corr \orcidID{0009-0001-9596-8115} \and
John W. Sheppard\orcidID{0000-0001-9487-5622}} 
\authorrunning{G. Popoola and J.W. Sheppard}


\institute{Gianforte School of Computing, Montana State University, Bozeman, MT, USA\\
{gideon.popoola@student.montana.edu, John.sheppard@montana.edu }
\tocauthor{Gideon Popoola, John W. Sheppard}
}
\maketitle              

\begin{abstract}
   Fairness in machine learning research has largely focused on outcome-oriented fairness criteria such as Equalized Odds, while comparatively less attention has been given to procedural-oriented fairness, which addresses how a model arrives at its predictions. Neglecting procedural fairness means it is possible for a model to generate different explanations for different protected groups, thereby eroding trust. In this work, we introduce ``Group Counterfactual Integrated Gradients (GCIG)", an in-processing regularization framework that enforces explanation invariance across groups, conditioned on the true label. For each input, GCIG computes explanations relative to multiple Group Conditional baselines and penalizes cross-group variation in these attributions during training. GCIG formalizes procedural fairness as Group Counterfactual explanation stability and complements existing fairness objectives that constrain predictions alone. We compared GCIG empirically against six state-of-the-art methods, and the results show that GCIG substantially reduces cross-group explanation disparity while maintaining competitive predictive performance and accuracy-fairness trade-offs. Our results also show that aligning model reasoning across groups offers a principled and practical avenue for advancing fairness beyond outcome parity.
\end{abstract}

\section{Introduction}
The main objective of fairness-aware machine learning is to ensure that model predictions do not systematically disadvantage or favor protected groups \cite{le2022survey}. Therefore, many research works have focused on \textit{outcome-oriented} metrics, such as statistical parity, equal opportunity, and equalized odds \cite{hardt2016equality}, which minimize the dependency between model predictions and protected attributes at the outcome level. Among these, Equalized Odds (EO) is particularly common, requiring that predictions be conditionally independent of protected attributes given the true label. When satisfied, EO ensures that error rates are balanced across groups, providing a strong notion of outcome-oriented fairness.

Unfortunately, outcome-oriented fairness guarantees alone do not fully characterize decision-making. Two models may achieve similar EO metrics while relying on fundamentally different decision processes across groups \cite{germino2025explanation}. 
Such discrepancies do not manifest in outcome-oriented fairness metrics; however, they reflect differences in \textit{how} decisions are reached rather than \textit{what} decisions are produced. This distinction has motivated growing interest in procedural fairness, which concerns the invariance of the decision-making process itself \cite{balagopalan2022road}.
 

Procedural fairness concerns whether a decision-making process applies consistent reasoning standards across different protected groups \cite{zhao2023fairness}. One way of measuring procedural bias is through feature-attribution, which decomposes model predictions into contributions from each feature, thus exposing the factors that drive model behavior on a per-instance basis \cite{dai2022fairness}. If two individuals with the same outcome are evaluated using substantially different feature attributions solely due to group membership, the model may be said to apply different reasoning processes across groups \cite{deck2024critical}. Therefore, the model would be procedurally biased towards one group.

Despite this potential, most existing uses of explanations in fairness contexts focus on pre-processing rather than in-processing \cite{wang2024procedural}. Explanations are typically computed after training to audit models for disparate treatment. While such analyses can reveal procedural disparities, they do not prevent them from arising in the first place. Moreover, explanation-based diagnostics are often applied at the individual level, making it difficult to detect systematic, group-level differences in model reasoning. Bridging this gap requires integrating explanation-based constraints directly into the learning process so that models are encouraged to learn not only fair outcomes but also consistent explanatory behavior across protected groups.


Overall, existing fairness methods lack mechanisms to enforce procedural fairness, while existing explanation methods lack training-time interventions aligned with fairness goals. This gap motivates the need for approaches that explicitly regulate explanations during training, ensuring that models not only achieve outcome fairness but also employ consistent reasoning across protected groups. In this work, we address this gap by proposing a training-time framework that enforces group counterfactual explanation invariance.


Our main idea is to assess and enforce this explanation invariance through group counterfactual explanations. 
Specifically, we seek to use informatiom from the question, \textit{how would the model’s explanation change if the same individual were evaluated relative to a different group context?}, to guide training. 
Large discrepancies indicate that the model’s reasoning depends on group-specific patterns rather than solely on the individual’s features. Building on this insight, we propose \textbf{Group Counterfactual Integrated Gradients (GCIG)}, a training-time regularization framework that minimizes cross-group variation in explanations. GCIG penalizes differences in Integrated Gradients computed relative to group conditional baselines, encouraging the model to produce explanations that are stable across group contexts. 

The main contributions from this work are threefold:
\begin{enumerate}
    \item We formalize procedural fairness as group counterfactual explanation invariance, a criterion requiring that model explanations remain consistent across protected groups conditional on the true label. 
    \item We propose Group Counterfactual Integrated Gradients (GCIG), an in-processing regularization approach that minimizes cross-group variation in feature attributions computed relative to Group Conditional baselines. 
    \item 
    We demonstrate across multiple benchmarks that GCIG reduces explanation disparity while maintaining competitive predictive performance and accuracy–fairness trade-offs relative to established fairness baselines.
\end{enumerate}

\section{Related Work}
Outcome fairness aims to ensure that a model’s predictions satisfy statistical parity constraints across protected groups at the population level \cite{roy2015impact}. Common criteria include statistical parity \cite{dwork2012fairness}, which requires predictions to be independent of protected attributes, and Equalized Odds (EO) \cite{hardt2016equality}, which requires predictions to be conditionally independent of protected attributes given the true label. Several works have developed algorithmic strategies to enforce outcome fairness. Pre-processing approaches modify the training data to reduce group disparities before modeling, for example, through reweighting or sampling \cite{popoola2024investigating}. In-processing methods incorporate fairness constraints directly into the learning objective, often via constrained optimization or adversarial training \cite{agarwal2018reductions}, while post-processing methods adjust predictions at inference time to satisfy fairness criteria without retraining the underlying model \cite{tifrea2023frappe}. Among all the approaches, in-processing methods have been shown to be the most effective method in mitigating bias \cite{wan2023processing}.


Model explainability seeks to make the behavior of complex model systems interpretable to humans. Among the most widely adopted approaches are feature-attribution methods
\cite{bodria2023benchmarking}. A popular class of feature attribution methods is based on gradients, including integrated gradient and their variants \cite{niebur2007saliency}. Integrated Gradients (IG) addresses limitations of raw gradients by accumulating gradients along a path from a baseline to the input of interest
\cite{sundararajan2017axiomatic}. Related methods, including Layer-wise Relevance Propagation (LRP) \cite{binder2016layer} and DeepLIFT \cite{li2021deep}, propagate relevance scores through the network structure to attribute predictions to input features. While these approaches differ in their computational mechanisms, they share the common goal of characterizing the model’s local decision logic. 

Recent work has begun to explore the use of model explanations to assess procedural fairness. Several research investigations examine explanation fidelity and stability, determining whether explanations remain consistent under perturbations of protected attributes \cite{zhao2023fairness}. 
Stability-based analyses have been applied to detect disparate treatment and to audit models for procedural inconsistencies that are not captured by outcome-based metrics \cite{wang2024procedural}. Some works also compare average feature attributions across groups to identify disparities in feature usage \cite{wang2024achieving}. 
While informative, most explanation-based fairness approaches are fundamentally diagnostic (i.e they evaluate trained models to identify explanation disparities but do not intervene during training to prevent such disparities from arising) \cite{leblanc2023relationship}. 
In contrast, our approach integrates explanation-level constraints directly into the learning objective and enforces explanation invariance conditional on the label, thereby aligning procedural fairness with EO. 

Procedural fairness has been studied in machine learning under several related but distinct frameworks that move beyond purely outcome criteria \cite{wang2024procedural}. Closely related to procedural fairness are counterfactual fairness frameworks \cite{kusner2017counterfactual} \cite{ma2023learning}, which evaluate whether a prediction would remain unchanged under hypothetical interventions on protected attributes within a causal model. 
Unfortunately, such methods do not directly address how to 
impose constraints on explanation-level behavior or ensure that the features driving predictions are consistent across protected groups. 

\section{Methodology}
\subsection{ Problem Setup and Notation}

We consider a supervised binary classification setting with protected attributes. Let $\mathcal{D} = \{(x_i, y_i, a_i)\}_{i=1}^n$ denote a dataset of $n$ instances, where $(x_i \in \mathcal{X} \subseteq \mathbb{R}^p)$ is a $p$-dimensional feature vector, $y_i \in \{0,1\}$ is the true label, and $a_i \in \mathcal{A}$ is a protected attribute. For simplicity, we limit the protected attribute to be binary, i.e., $\mathcal{A} = \{0,1\}$; however, GCIG can be extended in a straightforward manner to multiple protected categories.

Let $(X, Y, A)$ be sets of random variables distributed according to an unknown joint distribution $P(X,Y,A)$. 
Let $P_{y,g}(X) = P(X \mid Y=y, A=g)$ be the conditional feature distribution for label $y$ and protected group $g$.
We consider a 
prediction model $f_\theta : \mathcal{X} \to \mathbb{R}$, where $(f_\theta(x))$ produces a real-valued logit score. 

EO is a widely used group fairness criterion that requires predictions to be conditionally independent of the protected attribute given the true label. That is,
$\forall y \in \lbrace 0,1\rbrace)$ and $\forall g,g' \in \mathcal{A}$,
\[
P(\hat{Y}=1 \mid Y=y, A=g) = P(\hat{Y}=1 \mid Y=y, A=g').
\]
In practice, EO gaps are quantified using disparities in true positive and false positive rates:
\[
\Delta_{\mathrm{TPR}} = \big| \mathrm{TPR}_{0} - \mathrm{TPR}_{1} \big|,
\quad
\Delta_{\mathrm{FPR}} = \big| \mathrm{FPR}_{0} - \mathrm{FPR}_{1} \big|,
\]
where
\[
\mathrm{TPR}_g = P(\hat{Y}=1 \mid Y=1, A=g), 
\quad\mathrm{FPR}_g = P(\hat{Y}=1 \mid Y=0, A=g).
\]
A common scalar summary of EO violation is $\Delta_{\mathrm{EO}} = \Delta_{\mathrm{TPR}} + \Delta_{\mathrm{FPR}}$.

\subsection{Integrated Gradient}\label{IG}
Integrated Gradients (IG) provide a model explanation method that explains the predictions of differentiable models by attributing the change in the model output between a reference input and the input of interest to individual features. IG addresses limitations of raw gradients, such as saturation and noise, while satisfying axioms that formalize desirable properties of explanations.

Let $f_\theta$
be a differentiable model that outputs a logit score, and let $x, x' \in \mathcal{X}$ denote an input and a baseline (reference) input, respectively. The Integrated Gradient attribution for feature $j \in \lbrace 1,\dots,p\rbrace$ is defined as
\[
\mathrm{IG}_j(x; x') =(x_j - x'_j) \int_{0}^{1}
\frac{\partial f_\theta\big(x' + \alpha (x - x')\big)}{\partial x_j}
d\alpha.
\]
In vector form, the attribution is written as
\[
\mathrm{IG}(x; x') = (x - x') \odot \int_{0}^{1}
\nabla_x f_\theta\big(x' + \alpha (x - x')\big) d\alpha
\in \mathbb{R}^p \]
where $\odot$ denotes element-wise multiplication.
In practice, the integral is approximated using a Riemann sum with $T$ steps
\[
\mathrm{IG}(x; x') \approx (x - x') \odot \frac{1}{T}
\sum_{t=1}^{T}
\nabla_x f_\theta\left(x' + \frac{t}{T}(x - x')\right).
\]
A key property of Integrated Gradients is \textit{completeness}, which ensures that the total attribution equals the difference in model outputs between the input and the baseline:
\[
\sum_{j=1}^{p} \mathrm{IG}_j(x; x') = f_\theta(x) - f_\theta(x')
\]

Throughout this work, IG is computed with respect to the model’s logit output rather than the predicted probability $(\sigma(f_\theta(x)))$. This choice avoids gradient saturation caused by the sigmoid function and yields more stable, informative attributions during training.


\subsection{ Group Conditional Baselines}
In fairness settings, different protected groups can exhibit distinct feature distributions, and a single global baseline may obscure or distort group-specific context. To model this context, we introduce group conditional baselines to encode typical feature profiles for each protected group and label.

Specifically, for each label $y \in \{0,1\}$ and protected group $(g \in \mathcal{A})$, the group conditional baseline is defined as
\[
\mathbf{b}_{y,g} =
\mathbb{E}_{X \sim P(X \mid Y=y, A=g)}[X] \in \mathbb{R}^p.
\]
where $\mathbf{b}_{y,g}$ represents the average feature vector of individuals from group $g$ who share the same true label $y$. 
In practice, we can estimate $\mathbf{b}_{y,g}$ as
\[
\hat{\mathbf{b}}_{y,g} =
\frac{1}{|\mathcal{I}_{y,g}|}
\sum_{i \in \mathcal{I}_{y,g}} x_i,
\quad
\mathcal{I}_{y,g} := \{i : y_i=y, a_i=g\}.
\]
However, when used during stochastic training, such estimates can be unstable due to small minibatches or class imbalance. To address this issue, we maintain exponentially moving average (EMA) estimates of the baselines:
\[
\mathbf{b}_{y,g}^{(t)} \leftarrow
(1-\gamma)\mathbf{b}_{y,g}^{(t-1)}
+
\gamma\hat{\mathbf{b}}_{y,g}^{(t)},
\]
where $(\hat{\mathbf{b}}_{y,g}^{(t)})$ is the minibatch estimate at iteration ($t$), and $\gamma \in (0,1)$ is a smoothing parameter. EMA estimation provides stable, low-variance baseline estimates while adapting gradually as training progresses.

Group conditional baselines serve as contextual reference points rather than individual-specific counterfactuals. When IG is computed with respect to $(\mathbf{b}_{y,g})$, the resulting attribution answers the question: \textit{Which features distinguish this individual from a typical member of group $g$ with the same outcome?} Crucially, these baselines are treated as fixed reference points during gradient computation; gradients are not propagated through $(\mathbf{b}_{y,g})$. This design choice prevents degenerate solutions in which baselines adapt to minimize attribution disparity and ensures that explanation invariance is enforced through the model parameters rather than through shifting reference points. By making group context explicit through $(\mathbf{b}_{y,g})$, we enable controlled, counterfactual comparisons of explanations across protected groups.

\subsection{Group Counterfactual Integrated Gradients}
We now formalize procedural fairness via group counterfactual explanations and introduce the proposed explanation-disparity measure that underpins GCIG.
\subsubsection{Group Counterfactual Explanations} Given an input instance $x \in \mathcal{X}$ with true label $y \in \{0,1\}$, we compute IG with respect to each group conditional baseline defined above. For each protected group $g \in \mathcal{A}$, the Group Counterfactual Integrated Gradients attribution is defined as
\[
\mathrm{IG}^{(g)}(x; y)
=
\mathrm{IG}\big(x, \mathbf{b}_{y,g}\big)
\in \mathbb{R}^p.
\]
Each attribution vector $ \mathrm{IG}^{(g)}(x; y)$ represents the contribution of individual features to the model’s prediction for $x$, measured relative to the typical feature profile of group $g$ among individuals with the same label $y$. Importantly, the input $x$ is held fixed across groups; only the reference context (baselines) varies. This construction enables a controlled, counterfactual comparison of explanations across protected groups.
$\mathrm{IG}^{(g)}(x; y)$ can be explained as 
\textit{How does the model explain this prediction when evaluated relative to the feature distribution of group $g$?} If the model’s reasoning is invariant to group context, these explanations should be similar across $g$.

\subsubsection{Normalized Attributions}
Direct comparison of attributions across groups can be dominated by scale effects arising from feature magnitude or gradient norm. To focus on relative feature importance, we normalize each attribution vector. Let $(||\cdot||_q)$ denote an $(\ell_q)$ norm. The normalized attribution is defined as
\[
\widetilde{\mathrm{IG}}^{(g)}(x; y) =
\frac{\mathrm{IG}^{(g)}(x; y)}
{||\mathrm{IG}^{(g)}(x; y)||_q + \varepsilon},
\]
where $(\varepsilon > 0)$ ensures numerical stability. Unless otherwise specified, we use $q=2$. Normalization ensures that explanation comparisons reflect differences in feature reliance patterns rather than differences in overall attribution magnitude.

 \subsubsection{Explanation Disparity Across Groups}
We quantify procedural bias by measuring the variation in normalized attributions across group contexts. For a fixed input $x$ with label $y$, group counterfactual explanation disparity is defined as
\[
\mathbf{v}(x; y) =
\operatorname{Var}_{g \in \mathcal{A}}
\left[\widetilde{\mathrm{IG}}^{(g)}(x; y)\right]\in \mathbb{R}^p,
\]
where the variance is computed element-wise across groups. To obtain a scalar measure of disparity, we aggregate across features:
\[
V(x; y) =
\left| | \mathbf{v}(x; y)| \right|_2 =
\sqrt{\sum_{j=1}^{p} v_j(x; y)}.
\]
In the common case of a binary protected attribute $(|\mathcal{A}| = 2)$, the variance reduces to a scaled squared difference, corresponding to the distance between explanations under different group contexts:
\[
V(x; y) \propto
\left \| \widetilde{\mathrm{IG}}^{(0)}(x; y)-
\widetilde{\mathrm{IG}}^{(1)}(x; y)\right \|_2,
\]

The quantity $V(x; y)$ measures how sensitive the model’s explanation for $x$ is to the group context used as a reference. Low values indicate that the model relies on similar explanatory mechanisms across groups, while high values indicate group-dependent reasoning. Importantly, explanation disparity is defined conditional on the true label. This design ensures that explanations are compared only among instances that share the same outcome. 

\subsection{Training Objective}
We now integrate group counterfactual explanation disparity into a training objective that aligns procedural fairness with Equalized Odds. Our goal is to learn model parameters $\theta$ that achieve strong predictive performance while minimizing group-dependent variation in explanations conditioned on the true label.

For a fixed label $y \in \{0,1\}$, we define the expected group counterfactual explanation disparity as
\[
\mathcal{L}_{\mathrm{GCIG}}^{(y)}(\theta) =
\mathbb{E}_{X \sim P(X \mid Y=y)}
\left[V(X; y) \right]
\]
where $V(X; y)$ is the instance-level explanation disparity. This explanation sensitivity to group context among individuals with the same outcome.

To obtain a single scalar objective, we aggregate across labels:
\[
\mathcal{L}_{\mathrm{GCIG}}(\theta) =
\sum_{y \in \{0,1\}} \pi_y 
\mathcal{L}_{\mathrm{GCIG}}^{(y)}(\theta),
\]
where $\pi_y = P(Y=y)$ denotes label proportions. In practice, $\pi_y $ can be set to empirical label frequencies or fixed to $(\pi_0=\pi_1=0.5)$ for equal class emphasis. 


During training, expectations are approximated using mini-batches. Given a mini-batch $\mathcal{B} = {(x_i, y_i, a_i)}_{i=1}^{B}$, we compute the empirical GCIG loss as
\[
\widehat{\mathcal{L}}_{\mathrm{GCIG}}(\theta) =
\sum_{y \in \{0,1\}}
\frac{1}{|\mathcal{B}_y|}
\sum_{i \in \mathcal{B}_y}
V(x_i; y)
\]
where $\mathcal{B}_y = \{i \in \mathcal{B} : y_i = y\}$. This formulation ensures that each label contributes proportionally to the fairness objective, even under class imbalance. 

Finally, we combine performance, procedural fairness, and outcome-oriented fairness into a single objective function for training that we refer to as \textsc{FairX}. 
\[
\min_{\theta}\quad
\mathcal{L}_{\mathrm{total}}(\theta) =
\mathcal{L}_{\mathrm{pred}}(\theta) +
\lambda_{\mathrm{ig}} \mathcal{L}_{\mathrm{GCIG}}(\theta) +
\lambda_{\mathrm{fair}} \mathcal{L}_{\mathrm{fair}} 
\]
where $\mathcal{L}_{\mathrm{pred}}(\theta) $ is the standard binary cross-entropy loss, $\mathcal{L}_{\mathrm{GCIG}}(\theta)$ enforces explanation invariance across groups, $\lambda_{\mathrm{ig}} \ge 0$ controls the strength of procedural fairness regularization, $\mathcal{L}_{\mathrm{fair}}$ is outcome-oriented fairness (e.g., EO) and $\lambda_{\mathrm{fair}}$ controls the outcome-oriented fairness strength.
%
By jointly optimizing these objectives, our method ensures comprehensive fairness that goes beyond traditional fairness approaches. We show the complete steps of \textsc{FairX} in Algorithm \ref{alg:GCIG}.

\begin{algorithm}[t!]
\caption{\textsc{FairX} Regularized Training}
\label{alg:GCIG}
\begin{algorithmic}[1]
\Require Dataset $\mathcal{D}=\{(x_i,y_i,a_i)\}_{i=1}^n$, $y_i\in\{0,1\}$, $a_i\in\{0,1\}$; model $f_\theta$ (logit); IG steps $T$; batch size $B$; EMA rate $\gamma$; weights $\lambda_{\mathrm{ig}},\alpha$
\Ensure Trained parameters $\theta$ and EMA baselines $\{\mathbf{b}_{y,g}\}_{y\in\{0,1\},g\in\{0,1\}}$

\State Initialize $\theta$; initialize baselines $\mathbf{b}_{y,g}\in\mathbb{R}^p$ (e.g., from data means); set $\varepsilon>0$
\For{each epoch}
  \For{minibatch $\mathcal{B}$ of size $B$}
    \State \textbf{(EMA baselines)} $\hat{\mathbf{b}}_{y,g}\gets \frac{1}{|\mathcal{B}_{y,g}|}\sum_{i\in\mathcal{B}_{y,g}} x_i$; 
    $\mathbf{b}_{y,g}\leftarrow (1-\gamma)\mathbf{b}_{y,g}+\gamma \hat{\mathbf{b}}_{y,g}$ 
    \State \textbf{(Prediction loss)} $\mathcal{L}_{\mathrm{pred}}\gets \frac{1}{B}\sum_{i\in\mathcal{B}} \mathrm{BCE}\!\left(y_i,\sigma(f_\theta(x_i))\right)$
    \State $\mathcal{L}_{\mathrm{GCIG}}\gets 0$
    \For{$y\in\{0,1\}$}
      \State $\mathcal{L}_y\gets 0$
      \For{$i\in\mathcal{B}_y$} 
        \State \textbf{(Counterfactual IG)} $\mathrm{IG}^{(g)}_i \gets \mathrm{IG}(x_i;\mathbf{b}_{y,g})$ for $g\in\{0,1\}$ 
        \State \textbf{(Normalize)} $\widetilde{\mathrm{IG}}^{(g)}_i \gets \mathrm{IG}^{(g)}_i /(\|\mathrm{IG}^{(g)}_i\|_1+\varepsilon)$
        \State \textbf{(Disparity)} $V_i \gets \left\|\widetilde{\mathrm{IG}}^{(0)}_i-\widetilde{\mathrm{IG}}^{(1)}_i\right\|_2$
        \State $\mathcal{L}_y \gets \mathcal{L}_y + V_i$
      \EndFor
      \State $\mathcal{L}_{\mathrm{GCIG}} \gets \mathcal{L}_{\mathrm{GCIG}} + \frac{1}{|\mathcal{B}_y|}\mathcal{L}_y$ 
    \EndFor
    \State \textbf{(Update)} $\theta \leftarrow \theta - \eta \nabla_\theta\!\left(\mathcal{L}_{\mathrm{pred}}+\lambda_{\mathrm{ig}}\mathcal{L}_{\mathrm{GCIG}}
    + \lambda_{\mathrm{fair}} \mathcal{L}_{\mathrm{fair}} \right)$
  \EndFor
\EndFor
\end{algorithmic}
\end{algorithm}

\subsection{Computational Complexity}
The \textsc{FairX} algorithms is implemented as a training-time regularization procedure that augments standard empirical risk minimization with an explanation invariance loss. The overall training algorithm follows the standard mini-batch stochastic gradient descent pipeline with the addition of Group Counterfactual Integrated Gradients computation at each update. Since the protected attribute is binary in this work $(|\mathcal{A}| = 2)$, the algorithm simplifies considerably relative to the general multi-group formulation.

Let $B$ denote the mini-batch size, $p$ the number of input features, $T$ the number of integration steps used to approximate integrated gradients, and $C$ the cost of a single forward-backward pass of the model. Standard supervised training incurs a cost of $O(B \cdot C)$ per iteration. Computing integrated gradients for a single input with one baseline requires $(O(T \cdot C))$ operations. In GCIG, explanations are computed for each group conditional baseline. For $|\mathcal{A}| = 2$, the additional cost per batch is therefore
\[
O(B \cdot T \cdot C).
\]

To mitigate this overhead, we vectorize IG computation across group baselines, allowing all group counterfactual explanations for a given input to be computed in a single forward-backward pass per integration step. With this vectorization, the amortized complexity remains $(O(B \cdot T \cdot C))$, independent of the number of groups. In practice, we use small values of (T) (typically $T \in {8, 16}$) during training, which provides a favorable trade-off between computational cost and attribution accuracy. As a result, GCIG training is approximately ($T$)-times slower than standard training, but remains tractable for tabular and medium-scale datasets.

\section{Experimental Settings}
In this section, we seek to answer the following research questions:
\begin{enumerate}
    \item RQ1: Does \textsc{FairX} reduce group-dependent explanation disparity while maintaining predictive utility?
    \item RQ2: To what extent are outcome-oriented fairness and procedural fairness correlated across models and datasets?
    \item RQ3: How sensitive is GCIG to design choices?
\end{enumerate}

\subsection{Data}
\begin{table}[t]
\centering
\caption{Datasets used in the experimental evaluation}\label{data}
\begin{tabular}{|c|c|c|c|} 
\hline
Data           & Target                         & Samples & Features \\ \hline
Adult          & Income exceeds \$50k           & 4522    & 32       \\ \hline
German Credit  & Credit Worthiness              & 1000    & 30       \\ \hline
COMPAS         & Likelihood of Recidivism       & 6172    & 18       \\ \hline
Bank Marketing & Deposit subscription & 45211   & 21       \\ \hline
\end{tabular}
\end{table}

We use four publicly available state-of-the-art fairness datasets pulled from the UCI ML repository to evaluate our algorithm and baselines \cite{asuncion2007uci}. 
Continuous features were standardized to zero mean and unit variance, categorical variables were one-hot encoded, and each dataset was split into train/validation/test sets using a 60/20/20 split, stratified by both label and protected attribute. More details regarding the datasets can be found in Table \ref{data}

\subsection{Baselines}
To evaluate textsc{FairX}, we compared against established fairness interventions spanning unconstrained learning, pre-processing, post-processing, and in-processing approaches. Unconstrained ERM trains the same MLP architecture with binary cross-entropy on the training data, without fairness constraints. Post-processing by Hardt \cite{hardt2016equality} enforces equalized odds by learning group-specific decision thresholds on top of a base classifier, producing predictions that satisfy EO constraints up to estimation error. The Reductions method by Agrawal \cite{agarwal2018reductions} optimizes a cost-sensitive classification mixture via exponentiated gradients to satisfy equalized odds constraints, implemented using Fairlearn’s reductions framework. The Adversarial method \cite{zhang2018mitigating} trains a predictor jointly with an adversary that tries to infer the protected attribute from the predictor’s outputs, while the predictor learns representations that minimize prediction loss and reduce sensitive leakage. Disparate Impact Remover (DIR) \cite{feldman2015certifying} is a pre-processing method that reduces disparate impact by repairing feature distributions to diminish their dependence on the protected attribute while preserving as much rank/utility information as possible. Additionally, we include a Lagrangian primal-dual baseline that directly penalizes differentiable EO violations via dual updates on TPR/FPR gaps \cite{cotter2019two}. All methods are evaluated using the same train/test splits and standardized feature preprocessing. For methods requiring a base classifier (e.g., Hardt post-processing), we train the base model using the same architecture family as the unconstrained baseline. We perform grid search hyperparameter tuning for all methods separately, and we use the best parameters to run the experiments using 5-fold cross validation.

\subsection{Results}

To address RQ1, Table \ref{tab:main_results} summarizes predictive performance (F1), outcome fairness (EO gap), and procedural fairness (GCIG) across four datasets. Results are reported as mean $\pm$ standard deviation over 5-fold cross-validation. Across all datasets, \textsc{FairX} consistently reduces GCIG relative to the unconstrained baseline and all outcome-focused baselines. On German Credit, GCIG decreases from 0.190 (unconstrained) to 0.066. On COMPAS, GCIG decreases from 0.193 to 0.034. On Adult Income and Bank Marketing, similar reductions are observed $(0.191 \to 0.054$ and $0.169 \to 0.057$, respectively). These reductions indicate that explicitly penalizing group-conditional explanation variance substantially improves explanation alignment across protected groups.
Importantly, these procedural improvements do not result in systematic degradation in predictive performance. \textsc{FairX} achieves the highest F1 on German Credit (0.833) and matches the best-performing methods on COMPAS (0.967). On Adult and Bank, F1 remains within a small margin of the strongest baseline. This suggests that encouraging explanation invariance need not compromise utility in the evaluated settings.

Regarding outcome fairness, \textsc{FairX} achieves the lowest EO gap on German Credit (0.120) and remains competitive on the other datasets. On COMPAS, several methods including \textsc{FairX} achieve identical EO gaps (0.023), yet their GCIG values differ substantially. This shows that models satisfying similar outcome fairness constraints may exhibit markedly different explanation behavior.
Thus the results indicate that procedural fairness captures aspects of model behavior not reflected in outcome-based metrics alone, and that these aspects can be improved without degrading predictive performance.

\begin{table*}[t]
\centering
\caption{Performance comparison across four benchmark datasets. Results show mean $\pm$ std, where lower is better for EO Gap and GCIG and higher is better for F1. \textbf{Bold} indicate best performance}
\label{tab:main_results}
\resizebox{\textwidth}{!}{%
\begin{tabular}{llccccccc}
\toprule
\textbf{Dataset} & \textbf{Metric} & \textbf{Unconstrained} & \textbf{DIR} & \textbf{Hardt} & \textbf{Agarwal} & \textbf{Lagrangian} & \textbf{Adversarial} & \textbf{\textsc{FairX}} \\
\midrule

\multirow{3}{*}{German Credit}
& EO Gap  & $0.333 \pm 0.125$ & $0.300 \pm 0.160$ & $0.294 \pm 0.099$ & $0.160 \pm 0.132$ & $0.275 \pm 0.097$ & $0.217 \pm 0.127$ & $\mathbf{0.120 \pm 0.088}$ \\
& F1      & $0.830 \pm 0.014$ & $0.827 \pm 0.008$ & $0.819 \pm 0.023$ & $0.825 \pm 0.014$ & $0.831 \pm 0.015$ & $0.827 \pm 0.027$ & $\mathbf{0.833 \pm 0.008}$ \\
& GCIG  & $0.190 \pm 0.029$ & $0.202 \pm 0.036$ & $0.210 \pm 0.022$ & $0.213 \pm 0.008$ & $0.188 \pm 0.026$ & $0.152 \pm 0.015$ & $\mathbf{0.066 \pm 0.006}$ \\
\midrule

\multirow{3}{*}{COMPAS}
& EO Gap  & $\mathbf{0.023 \pm 0.016}$ & $0.071 \pm 0.067$ & $\mathbf{0.023 \pm 0.016}$ & $0.031 \pm 0.020$ & $\mathbf{0.023 \pm 0.016}$ & $\mathbf{0.023 \pm 0.016}$ & $\mathbf{0.023 \pm 0.016}$ \\
& F1      & $\mathbf{0.967 \pm 0.003}$ & $0.948 \pm 0.028$ & $\mathbf{0.967 \pm 0.003}$ & $0.959 \pm 0.004$ & $\mathbf{0.967 \pm 0.003}$ & $\mathbf{0.967 \pm 0.003}$ & $\mathbf{0.967 \pm 0.003}$ \\
& GCIG  & $0.193 \pm 0.018$ & $0.147 \pm 0.010$ & $0.185 \pm 0.038$ & $0.190 \pm 0.008$ & $0.192 \pm 0.027$ & $0.163 \pm 0.014$ & $\mathbf{0.034 \pm 0.006}$ \\
\midrule

\multirow{3}{*}{Adult Income}
& EO Gap  & $0.142 \pm 0.050$ & $0.186 \pm 0.097$ & $0.175 \pm 0.049$ & $\mathbf{0.061 \pm 0.039}$ & $0.138 \pm 0.050$ & $0.147 \pm 0.060$ & $0.141 \pm 0.059$ \\
& F1      & $0.669 \pm 0.013$ & $0.656 \pm 0.015$ & $0.667 \pm 0.015$ & $0.627 \pm 0.013$ & $\mathbf{0.680 \pm 0.012}$ & $0.657 \pm 0.014$ & $0.676 \pm 0.017$ \\
& GCIG  & $0.191 \pm 0.010$ & $0.255 \pm 0.025$ & $0.248 \pm 0.003$ & $0.245 \pm 0.012$ & $0.191 \pm 0.009$ & $0.190 \pm 0.009$ & $\mathbf{0.054 \pm 0.005}$ \\
\midrule

\multirow{3}{*}{Bank Marketing}
& EO Gap  & $0.076 \pm 0.043$ & $0.132 \pm 0.062$ & $0.052 \pm 0.033$ & $\mathbf{0.050 \pm 0.032}$ & $0.102 \pm 0.042$ & $0.067 \pm 0.043$ & $0.062 \pm 0.019$ \\
& F1      & $0.538 \pm 0.032$ & $0.507 \pm 0.036$ & $0.253 \pm 0.124$ & $0.449 \pm 0.022$ & $\mathbf{0.548 \pm 0.021}$ & $0.540 \pm 0.018$ & $0.534 \pm 0.014$ \\
& GCIG  & $0.169 \pm 0.021$ & $0.169 \pm 0.014$ & $0.116 \pm 0.009$ & $0.120 \pm 0.005$ & $0.176 \pm 0.008$ & $0.174 \pm 0.014$ & $\mathbf{0.057 \pm 0.009}$ \\

\bottomrule
\end{tabular}%
}
\end{table*}

\begin{table}[t]
\centering
\caption{Pairwise performance comparisons. For each metric, ``Win'' indicates datasets where the baseline outperforms \textsc{FairX}, ``Loss'' indicates datasets where \textsc{FairX} outperforms the baseline, and ``Champ'' indicates datasets where the method achieves the best overall score.}
\label{tab:pairwise_comparison}
\begin{tabular}{|l|ccc|ccc|ccc|}
\hline 
 & \multicolumn{3}{c|}{\textbf{Proc. Fair}} & \multicolumn{3}{c|}{\textbf{Out. Fair}} & \multicolumn{3}{c|}{\textbf{Utility}}\tabularnewline
\cline{2-10}
 & \multicolumn{3}{c|}{GCIG ($\downarrow$)} & \multicolumn{3}{c|}{EO Gap ($\downarrow$)} & \multicolumn{3}{c|}{F1 ($\uparrow$)}\tabularnewline
\cline{2-10}
\textbf{Method} & Win & Loss & Champ & Win & Loss & Champ & Win & Loss & Champ\tabularnewline
\hline 
Unconstrained & 0 & 4 & 0 & 0 & 3 & 1 & 1 & 2 & 1\tabularnewline
DIR & 0 & 4 & 0 & 0 & 4 & 0 & 0 & 4 & 0\tabularnewline
Hardt & 0 & 4 & 0 & 1 & 2 & 1 & 0 & 3 & 1\tabularnewline
Agarwal & 0 & 4 & 0 & 2 & 2 & 2 & 0 & 4 & 0\tabularnewline
Lagrangian & 0 & 4 & 0 & 1 & 2 & 1 & 2 & 1 & 3\tabularnewline
Adversarial & 0 & 4 & 0 & 0 & 3 & 1 & 1 & 2 & 1\tabularnewline
FairX & 4 & 0 & 4 & 2 & 2 & 1 & 2 & 2 & 1\tabularnewline
\hline 
\end{tabular}
%
\end{table}

Table \ref{tab:pairwise_comparison} presents pairwise performance comparisons across all datasets. GCIG achieves champion status on all 4 datasets for procedural fairness (4/4, 100\%), demonstrating superiority in explanation consistency. For outcome fairness and utility, GCIG achieves champion status on 2 out of 4 datasets, indicating competitive but not exclusive dominance. Methods specializing in single objectives, such as Lagrangian for utility (3 championships) or Agarwal for outcome fairness (2 championships), achieve narrow superiority at the cost of significantly worse procedural fairness (0 championships, 0 wins vs GCIG on all 4 datasets).

We next examine RQ2 which asks whether outcome-oriented fairness (Equalized Odds) and procedural fairness (GCIG) are statistically aligned in practice.
Using fold-level results from RQ1 (140 model instances), we compute correlations between EO gap and GCIG across all datasets and methods. Across all folds, the Pearson correlation is $r = 0.244$ ($p = 0.0037$), and the Spearman rank correlation is $\rho = 0.229$ ($p = 0.0064$). While statistically significant, the effect size is small, indicating weak alignment between outcome and procedural fairness.

To quantify explanatory power, we fit the linear model:
\[
\text{GCIG} = \beta_0 + \beta_1 \cdot \text{EO} + \epsilon.
\]
Equalized Odds explains only $R^2 = 0.059$ of the variance in procedural disparity, indicating that over $94\%$ of explanation variance remains unexplained by outcome fairness. Importantly, the relationship is dataset-dependent. Moderate correlation appears on German Credit ($r = 0.383$, $p = 0.0231$), but vanishes on Adult ($r = -0.018$, $p = 0.9181$) and COMPAS ($r = -0.117$, $p = 0.5027$), with weak alignment on Bank ($r = 0.263$, $p = 0.1277$). 

This instability suggests that outcome and procedural fairness capture partially distinct behavioral dimensions whose interaction depends on data characteristics. Partial correlation controlling for predictive performance (F1) remains virtually unchanged ($r = 0.244$, $p = 0.0037$), confirming that the observed relationship is not driven by model accuracy.
Thus, these findings indicate that satisfying Equalized Odds does not guarantee explanation-level alignment, supporting the need for explicit procedural constraints when explanation consistency across groups is desired.

\begin{figure}
    \centering
    \includegraphics[width=0.9\linewidth]{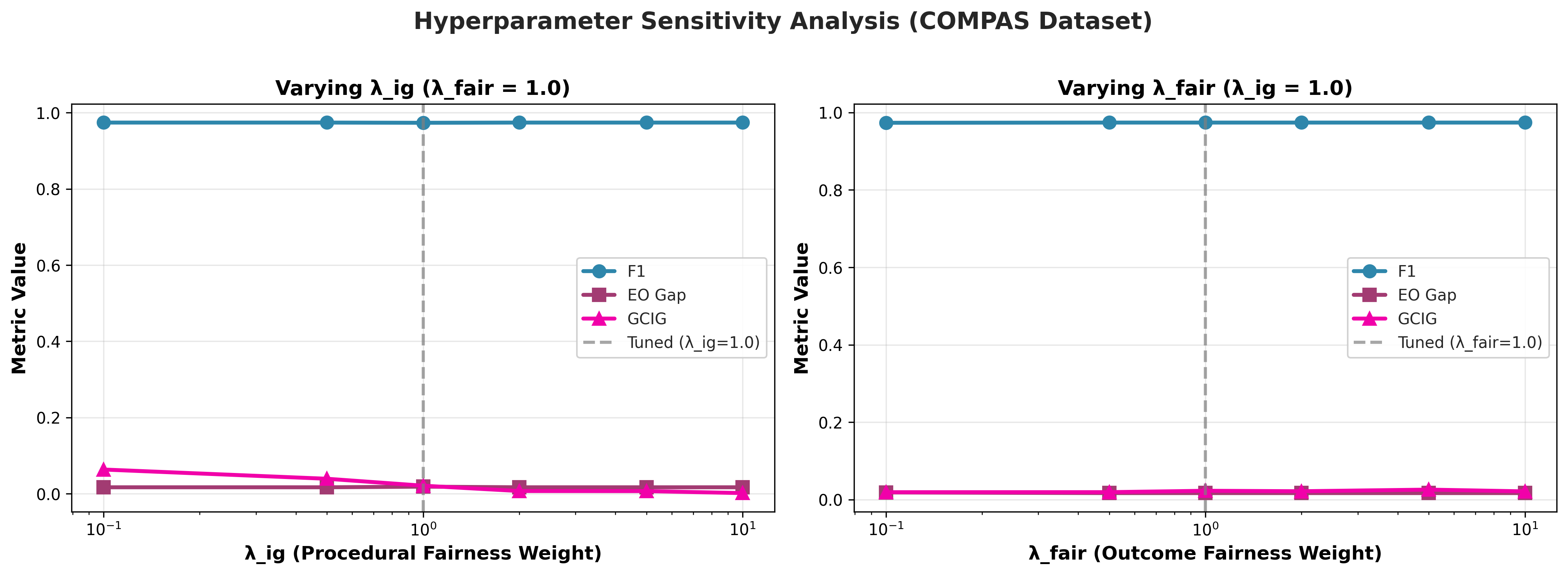} 
    \caption{Sensitivity analysis of $\lambda_{\mathrm{ig}}$ and $\lambda_{\mathrm{fair}}$ on COMPAS dataset} \label{sens}
    \label{fig:placeholder}
\end{figure}

To evaluate the robustness of our \textsc{FairX} to hyperparameter selection, we conducted a sensitivity analysis on the COMPAS dataset by independently varying the regularization weights $\lambda_{\mathrm{ig}}$ (procedural fairness) and $ \lambda_{\mathrm{fair}}$ (outcome fairness) across two orders of magnitude, from 0.1 to 10.0. For the first experiment, we fixed $\lambda_{\mathrm{fair}}$ at its tuned value (1.0) and tuned $\lambda_{\mathrm{ig}}$ across six values: {0.1, 0.5, 1.0, 2.0, 5.0, 10.0}. For the second experiment, we fixed $\lambda_{\mathrm{ig}}$ at 1.0 and varied $\lambda_{\mathrm{fair}}$ across the same range. 
The result in Figure \ref{sens} shows the robust performance of our method and it also consistent with prior observations that performance–fairness trade-offs may not always be inevitable \cite{popoola2024investigating}. These results also suggest that \textsc{FairX} does not require precise hyperparameter tuning to achieve stable performance.

\subsection{Ablation Study}
Finally, to address RQ3, we conducted an ablation study on COMPAS by training models with: (1) no fairness constraints (Prediction Only), (2) outcome fairness only (Prediction + EO), (3) procedural fairness only (Prediction + GCIG), and (4) both fairness components (Full Model). Results shown in Table \ref{tab:ablation} reveal that both components are necessary, adding outcome fairness alone worsens procedural fairness by 14\% (GCIG: 0.030 → 0.034), while adding procedural fairness alone provides modest 6\% improvement (GCIG: 0.028). Combining both, however, achieves 24\% improvement (GCIG: 0.023), substantially better than either component individually. This synergistic effect demonstrates that optimizing outcome fairness without considering explanation consistency degrades reasoning patterns, validating our dual-fairness framework. The Full Model maintains competitive utility (AUC: 0.979, -0.2\%) and identical outcome fairness (EO: 0.017), showing that joint optimization need not degrade individual objectives.


\begin{table}[t]
\centering
\caption{Ablation study showing the contribution of each objective component (COMPAS dataset). Percentage changes relative to Prediction Only baseline shown in parentheses for GCIG.}
\label{tab:ablation}
\begin{tabular}{lcccc}
\toprule
\textbf{Configuration} & \textbf{AUC} & \textbf{F1} & \textbf{EO Gap} & \textbf{GCIG} \\
\midrule
Prediction Only          & 0.981      & 0.974      & 0.017      & 0.030      \\
Prediction + EO          & 0.800      & 0.974      & 0.017      & 0.034 {\scriptsize (+14\%)} \\
Prediction + GCIG      & 0.980      & 0.974      & 0.017      & 0.028 {\scriptsize (-6\%)} \\
\textbf{Full Model}      & \textbf{0.979} & \textbf{0.974} & \textbf{0.017} & \textbf{0.023} {\scriptsize \textbf{(-24\%)}} \\
\bottomrule
\end{tabular}
\end{table}

\section{Discussion }
The results presented in Tables \ref{tab:main_results} and \ref{tab:pairwise_comparison} provide empirical evidence that outcome-oriented fairness metrics alone do not fully characterize differences in model behavior across protected groups. In several cases, methods achieving comparable Equalized Odds (EO) gaps exhibited substantially different levels of explanation disparity as measured by GCIG. For example, on the COMPAS dataset, multiple methods achieve identical EO gaps (0.023), yet their procedural fairness scores vary by a factor of more than five. This suggests that equalized error rates do not necessarily imply similarity in the underlying decision process.

Our findings indicate that explanation-level disparities constitute an independent dimension of fairness. Methods optimized exclusively for outcome criteria, such as Agarwal or Adversarial approaches, can achieve low EO gaps while still exhibiting substantial variation in feature attributions across protected groups. However, incorporating GCIG regularization reduces explanation disparity without materially degrading predictive performance. Across datasets, \textsc{FairX} consistently lowers procedural disparity while remaining competitive in both F1 and EO gap, with performance differences typically within a small margin of the strongest baseline for each metric.

Importantly, the results do not suggest that procedural and outcome fairness objectives are inherently conflicting. In our experiments, reducing GCIG did not systematically worsen EO gaps or predictive utility. Rather, the objectives appear complementary in many settings. However, this complementarity should not be interpreted as universal; it may depend on dataset characteristics, model class, and the relative weighting of objectives. Further theoretical and empirical analysis is required to characterize conditions under which joint optimization remains stable.


The ablation study supports the claim that both components of the objective contribute to procedural consistency. Optimizing outcome fairness alone does not improve explanation invariance and may modestly increase explanation disparity. In contrast, the inclusion of GCIG reduces explanation variance across groups, and the joint objective yields the strongest procedural improvement. Notably, these gains occur with minimal changes in predictive metrics, indicating that explanation alignment can be encouraged without significantly altering classification behavior.

Overall, our results suggest that explanation-based regularization provides a principled mechanism for complementing outcome-oriented fairness objectives. By integrating procedural considerations directly into training, \textsc{FairX} moves beyond post-hoc auditing and offers a unified framework for addressing multiple dimensions of fairness simultaneously.

\section{Conclusion}

We introduced Group Counterfactual Integrated Gradients (GCIG), a procedural fairness criterion that measures explanation disparity across protected groups conditional on the true label. By computing integrated gradients with respect to group conditional baselines and penalizing cross-group variation during training, we operationalize procedural fairness as explanation invariance. We further incorporated GCIG into a multi-objective training framework (\textsc{FairX}) that jointly optimizes predictive utility, outcome fairness, and procedural fairness. Empirical results across four benchmark datasets show that explicitly regularizing explanation disparity substantially reduces group-dependent variation in feature attributions while maintaining competitive predictive performance and outcome fairness. 

This work has several limitations. First, our experiments focus on binary protected attributes and tabular datasets; extending GCIG to multi-valued or intersectional attributes and to other modalities (e.g., text or images) requires further study.  Also, while GCIG is computationally tractable for moderate-scale problems, its reliance on integrated gradients introduces additional training overhead.

Future work will explore theoretical characterizations of explanation invariance, extensions to multi-group settings, and alternative attribution methods beyond integrated gradients. More broadly, we view this work as a step toward integrating procedural considerations into fairness-aware learning objectives, moving beyond post-hoc auditing toward training-time alignment of both predictions and explanations.
\bibliographystyle{splncs04}
\bibliography{aaai25}

\end{document}